\documentclass{article}
\usepackage[numbers]{natbib}
\usepackage {arxiv}
\usepackage{cite}

\usepackage{floatrow}
\usepackage{hyperref}
\usepackage{amsmath}
\usepackage{amssymb}
\usepackage{mathtools}
\usepackage{amsthm}
\usepackage{amsmath,amssymb,amsfonts}
\usepackage{algorithm}
\usepackage{algorithmic}
\usepackage{textcomp}
\usepackage{xcolor}
\usepackage{float}
\usepackage{booktabs}
\usepackage{multirow}
\usepackage{enumitem}
\usepackage{setspace}
\usepackage[]{graphicx}
\usepackage{array}
\usepackage{amsmath}
\usepackage{amssymb}
\usepackage{epsfig}
\usepackage{epstopdf}
\usepackage{subfigure}
\usepackage{comment}
\usepackage{tabularx}
\usepackage{threeparttable}
\usepackage{adjustbox}
\usepackage{rotating}
\usepackage{soul}
\usepackage{dsfont}
\usepackage{hyperref} 
\usepackage{lineno}
\usepackage{soul}

\def\BibTeX{{\rm B\kern-.05em{\sc i\kern-.025em b}\kern-.08em
    T\kern-.1667em\lower.7ex\hbox{E}\kern-.125emX}}

 \title{\LARGE \bf Imputation-free Learning of Tabular Data with Missing Values using Incremental Feature Partitions in Transformer}

\author{Manar D. Samad,~
       Kazi Fuad B. Akhter,~
        Shourav B. Rabbani, ~
        Ibna Kowsar\\
Department of Computer Science\\
Tennessee State University\\
Nashville, TN, USA\\
\texttt{msamad@tnstate.edu} \\
}

\begin{document}

\maketitle

\begin{abstract}
Tabular data sets with varying missing values are prepared for machine learning using an arbitrary imputation strategy. Synthetic values generated by imputation models often raise concerns regarding data quality and the reliability of data-driven outcomes. To address these concerns, this article proposes an imputation-free incremental attention learning (IFIAL) method for tabular data with missing values. A pair of attention masks is derived and retrofitted to a transformer to directly streamline tabular data without imputing or initializing missing values. The proposed method incrementally learns partitions of overlapping and fixed-size feature sets to enhance the performance of the transformer. The average classification performance rank order across 17 diverse tabular data sets highlights the superiority of IFIAL over 11 state-of-the-art learning methods with or without missing value imputations. Additional experiments corroborate the robustness of IFIAL to varying types and proportions of missing data, demonstrating its superiority over methods that rely on explicit imputations. A feature partition size equal to one-half the original feature space yields the best trade-off between computational efficiency and predictive performance. IFIAL is one of the first solutions that enables deep attention models to learn directly from tabular data, eliminating the need to impute missing values. The source code for this paper is available at~\footnote{\href{https://anonymous.4open.science/r/ifial-435B}%
{https://anonymous.4open.science/r/ifial-435B}}.

\end{abstract}


\section{Introduction}

Missing values, occurring at varying rates and types, are ubiquitous in health sciences research using electronic health record (EHR) data. For two reasons, handling missing values remains critical to ensure unbiased data and reliable data-driven results. First, the state-of-the-art machine learning (ML) methods that promise precision medicine are not designed to learn data with missing values. Second, missing values are arbitrarily handled by excluding samples and features or imputing data with model-generated synthetic values to enable machine learning. Sample exclusion biases medical data with sicker patients when their healthy counterparts are known to have more missing values~\citep{hubbard2021}. Feature exclusion due to high missing rates can remove potentially important predictors. Furthermore, many health scientists are skeptical of using medical data imputed with synthetic values generated by models, especially when ML models are hard to interpret. Imputed values can alter feature importance, data statistics, and correlations between features, dampening the predictive ability of data~\citep{Arifeen2022}. For example, multiple imputations using chained equations (MICE) is arguably the most popular method in health science to impute missing values. MICE estimates missing values using a chain of linear regression models for individual features with two major limitations. First, regression models are ineffective when features have limited or no relationships with each other. Second, the relationship between features is rarely linear in practice. In other words, data statistics and missing value patterns should guide the modeling and selection of an imputation algorithm. Clearly, there is no best imputation method for all data sets or missing value types~\citep{payrovnaziri2021,Ribeiro2021, samad2022}. All concerns surrounding imputation methods underscore the need for alternatives to imputation in learning data with missing values. This paper proposes Imputation-free Incremental Attention Learning (IFIAL) of tabular data with missing values. IFIAL can directly classify tabular data with missing values, eliminating biases and concerns surrounding data exclusions and missing value imputation. 

The remainder of this paper is organized as follows. Section~\ref{related_work} reviews the relevant literature on incremental feature learning and the learning of tabular data with or without missing values to motivate the contributions of our work. Section~\ref{methodology} introduces the proposed incremental learning and attention mask mechanism to develop the IFIAL framework. Section~\ref{experiments} discusses the experimental setup, data sets, missingness scenarios and baselines, and the evaluation method. Section~\ref{results} presents the performance of the proposed method and compares it with the state-of-the-art baselines, including robustness and ablation studies. Section~\ref{discussions} summarizes the key findings, provides insight into the results, and highlights the limitations of the proposed method. Finally, Section~\ref{conclusions} concludes the paper.
   
\subsection{Related work} \label{related_work}
The superiority of traditional ML in tabular data over deep learning is well documented in the literature~\citep{Borisov2022}. Feature incremental learning (FIL) is one of the cornerstones of deep learning (DL), which is not viable in traditional ML. Although highly successful with homogeneous image features~\citep{Li2022}, incremental learning is challenging for tabular data due to a heterogeneous feature space~\citep{Liu2023}. Some recent work has proposed FIL on tabular data~\citep{Liu2023, Ahamed2024mambatab, Kim2024rail, Wang2024Transtab}. A common approach to FIL trains multiple models separately to accommodate new features and aggregate model predictions through a weighting mechanism~\citep{Kim2024rail, Liu2023}. Alternatively, mamba~\citep{Ahamed2024mambatab} and the transformer-based method~\citep{Wang2024Transtab} use a single model to train on additional features incrementally as they become available. However, the efficacy of FIL has not been investigated in learning tabular data with missing values, especially in handling missing values.

The state-of-the-art methods for handling missing values are broadly categorized into traditional and deep imputations. Traditional methods include MICE~\citep{Lauren2021}, missForest~\citep{Stekhoven2011}, and MLP in multiple imputations~\citep{samad2022}). These methods train regression models to estimate and impute missing values, assuming the existence of relationships between features. Deep imputation methods include Generative Adversarial Imputation Nets (GAIN)~\citep{yoon2018gain}, diffusion-based imputation (Diffputer)~\citep{diffputer}, variational autoencoders~\citep{Hong2023}, and denoising autoencoders~\citep{Shang2017}. Sun et al. ~\citep{sun2023deep} report that GAIN has limited performance when missing values are at random (MAR) or not at random (MNAR), where MICE often outperforms GAIN. However, generative adversarial networks (GANs) can be unstable during optimization, particularly in terms of model convergence~\citep{hyperimpute}. Diffusion-based methods employ a simplified assumption about feature distributions (e.g., a Gaussian distribution), which may not be optimal for yielding the most accurate estimation of missing values. Chen et al.~\citep{rethinking_diff} report the superiority of GAIN over its diffusion counterparts, such as MissDiff~\citep{ouyang2023missdiff} and CSDI\_T~\citep{tabcsdi}. These data-centric assumptions and limitations of the imputation methods explain why no single imputation method is ideal for all data sets and missing value types. However, domain researchers have been reluctant to adopt DL methods for missing value imputation due to high computational costs, lack of trust, and interpretability~\citep{Bansal2021, Kamal2020}. Several DL methods have been proposed to learn data with missing values without using a standalone imputation step~\citep{Morvan}. However, similar methods first initialize missing values to subsequently learn a complete data representation during an end-to-end supervised learning task, such as classification and regression. Although the contribution of missing values to learning the complete data representation is masked out, these models do not distinguish between an observed zero value and zeros representing missing entries for a feature. Ideally, a DL framework should directly learn tabular data with missing values by streamlining only the observed values to address concerns about data quality and computational costs. In fact, decision tree-based ML models such as Random Forest and XGBoost~\citep{chen2016xgboost} use ``Missing Incorporated in Attributes" (MIA)\citep{twala2008mia} to directly learn data with missing values. However, traditional ML methods can be prone to overfitting high-dimensional feature spaces when incremental feature learning is not viable.

\begin{figure*}[t]
\centerline{\includegraphics[width=1.2\textwidth]{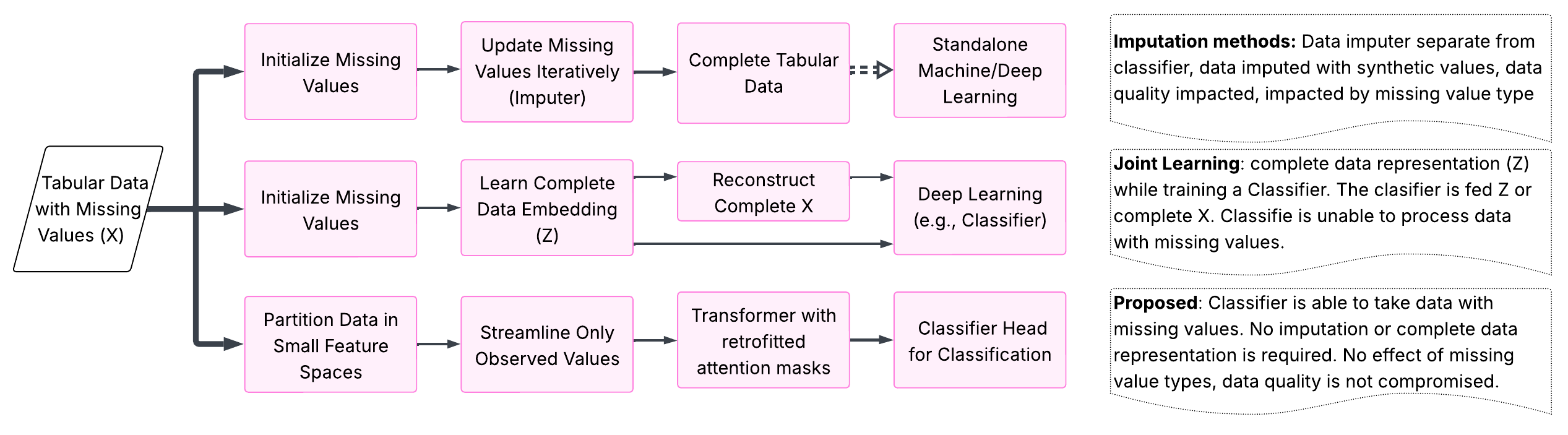}}
\caption{Three strategies for handling missing values in tabular data. Top: Standalone imputation of missing values before classification. Middle: Joint learning of complete data representation and classification. Bottom: (Proposed) - no imputation or initialization of missing values needed to streamline only the observed values for classification.}
\label{fig:contributions}
\end{figure*}

\subsection {Motivation and contributions}
\label{contributions}

Figure~\ref{fig:contributions} demonstrates current strategies for handling missing values compared to our proposed imputation-free approach. A standalone method for imputing missing values can compromise data quality when the statistics underlying the missing value type are unknown. The selection of an imputation method for a given data type or distribution is not trivial. Furthermore, there is no single best imputation method for all tabular data sets. Tabular data with mixed feature types require separate classifier and regressor models to impute numerical and categorical features, respectively. When the entire data set is used to build the imputation model, similar to~\citep{du2024remasker, Lall_Robinson_2022}, the test data fold in subsequent classification is likely to contain information leaked from the training data folds. Alternatively, the joint imputation and classification methods~\citep{KBS-Impute} impute data (X) to fit the classification targets (y), as X$\leftarrow$f(y), opposite to the premise of a classification task, y $\leftarrow$ f(X). Combining classification and imputation learning can leak information from class labels to data values (X), resulting in biased data for classification. Instead, a complete data representation or embedding (Z) can be learned from data (X) with missing values during supervised training of a classifier~\citep{Morvan}. Existing joint learning methods require the initialization of missing values to prepare a complete data representation for a classifier or regressor that would otherwise be unable to learn data with missing values directly. However, without complete data representation, tabular data would consist of features with varying missing rates, which can impact a classifier's ability to learn effectively from the relationships between features. In this context, sequential learning of data in small feature partitions, starting from the partition with the lowest missing rates, can mitigate the effect of varying missing rates and improve computational efficiency. Therefore, the goal of this paper is to enable a state-of-the-art deep attention learning classifier to directly and effectively learn tabular data with missing values in small feature partitions, bypassing the conventional need for data imputation, missing value initialization, and learning complete data representation. This article makes five key contributions to address the challenges in learning tabular data with missing values. 
\begin{itemize}
    \item We present one of the first studies to streamline tabular data with missing values in a deep attention learning framework without requiring imputation or even initialization of missing values. 
    \item The proposed method derives and retrofits two attention masks to a feature-tokenized transformer to exclude missing entries from attention scoring. \item We propose a new incremental learning strategy to mitigate the effect of varying missing rates in the feature space and allow efficient attention learning of O ($n^2$), which otherwise results in an out-of-memory problem in large feature spaces~\citep{rabbani2024}. 
    \item The proposed incremental learning partitions the data into small and fixed-size feature spaces, rather than gradually increasing feature sizes, allowing more efficient learning of a large-dimensional feature space. 
    \item The entire deep attention framework is designed to be computationally efficient and performancewise superior to state-of-the-art baselines, both with and without missing value imputations.  
\end{itemize}

\section {Method} 
\label{methodology}
This section presents the missing value problem, incremental learning, the attention mask mechanism, and the proposed algorithm combining the attention mask within an incremental learning framework. 

\subsection{Preliminaries} 

A tabular data set \( X \in \Re^{n \times d} = \{X_{obs}, X_{miss}\}\) with $r\%$ missing values is composed of observed ($X_{obs}$) and missing ($X_{miss}$) values. An imputation model first initializes $X_{miss}$ using mean or median values as ($\hat {X}_{miss}$). The model then iteratively refines the initial estimates of missing values as $\hat {X}_{miss}^{i+1}$ using $X_{obs}$ and $\hat {X}_{miss}^i$ estimates from the previous iteration. An imputation model makes the assumption that $X_{obs}$ is predictive of $X_{miss}$, which may not always be true depending on the predictive model, data statistics, and missing value types. This assumption can be entirely waived when a data set can be learned using $X_{obs}$ alone without estimating the $X_{miss}$ portion. A data set with missing values can be learned through FIL using $X_{obs}$ alone. FIL updates an ML model when additional features become available in a multiphase data collection scenario. An ML model trained on the data set $X_1 \in \Re^{N_1 \times d_1}$ can be updated by a data set collected in the second phase $X_2 \in \Re^{N_2 \times (d_1 + d_2)}$ with additional $d_2$ features, and so on. Without FIL, the conventional approach combines the two tabular data sets $X_1$ and $X_2$ to train the ML model from scratch, where samples of $X_1$ would have values missing for additional $d_2$ features. Unfortunately, traditional ML is not usually designed to learn directly from data with missing values. In this context, we repurpose FIL to learn data with missing values directly, eliminating the need for data imputation or exclusion, as discussed in subsequent sections. 

\begin{figure*}[t]
\centerline{\includegraphics[width=1.2\textwidth]{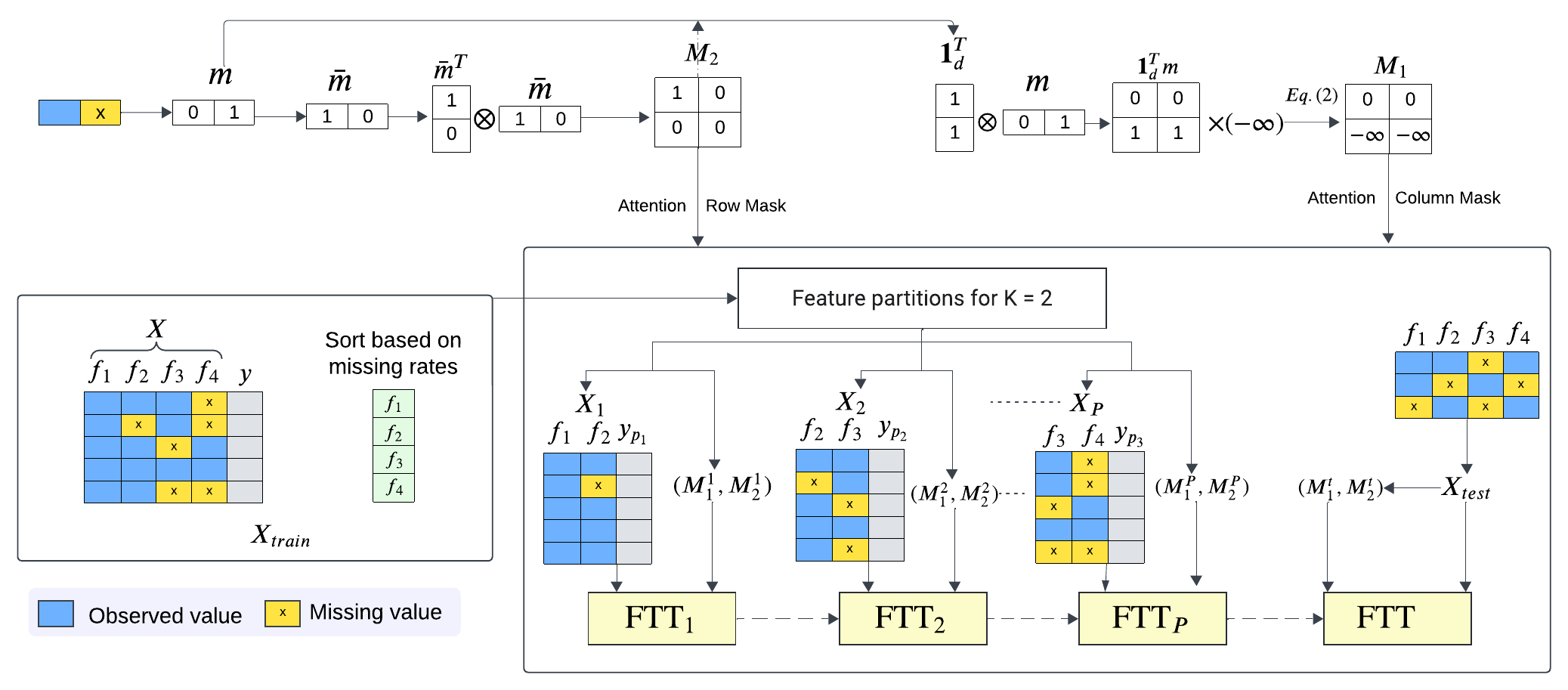}}
\caption{Imputation-free Incremental Attention Learning (IFIAL) algorithm uses $P$ fixed-sized overlapping feature partitions to train a Feature-Tokenized Transformer (FTT) incrementally. Attention masks: $M_1$ operates as the attention column mask and $M_2$ as the attention row mask to exclude missing feature values from attention scoring. 
}
\label{fig-chunking}
\end{figure*} 
  
\subsection{Proposed IFIAL algorithm}The proposed Imputation-free Incremental Attention Learning (IFIAL) completely avoids imputation or initialization of missing values $X_{miss}$. Suppose a labeled tabular data set with missing values \( D = \{X, y\} \) has \( n \) samples and $d$ features. First, we sort the features \( f_1, f_2, \dots, f_m \) in ascending order of missing rates. The sorted features are partitioned into subsets of \( k \) features with $ s = ceil (\frac{k}{2})$ overlaps, as shown in Figure \ref{fig-chunking}. The overlap between two consecutive feature subsets facilitates incremental learning using $P$ partitions of features, as follows.
\begin{equation}
\label{eq:p}
P = 1 + \lceil{\frac {d - k} { k - \lceil{k/2}\rceil}}\rceil
\end{equation}
These $P$ partitions are used to achieve incremental learning using a feature-tokenized transformer (FTT)~\citep{Wang2024Transtab,huang2020, schambach2023} as follows. An FTT classifier model ($FTT_1$) is trained using the first partition $ P_1 = X\{f_1, ..., f_k\}$ of $k$ features. The next incremental learning session uses the second partition $P_2 = X\{f_s, ..., f_{k+s}\}$ to train the previous $FTT_1$ model to $FTT_2$. Incremental learning continues until the $FTT_P$ model is obtained using the last feature partition, as illustrated in Figure~\ref{fig-chunking}. The trained $FTT_P$ model directly classifies a separate test data fold with missing values. Notably, FTT tokenizes features and uses a pre-trained language model to obtain embeddings for feature names and categorical features. In contrast, numerical values are passed through a linear projection layer to obtain corresponding embeddings. The tabular data representation is initialized by aggregating the embeddings of feature names and numerical and categorical feature values, which are subsequently fine-tuned for classification using a gated transformer. The proposed IFIAL algorithm is presented in Algorithm 1.  Missing values ($X^{miss}$) of X are excluded from attention scoring using attention masks designed as follows.

\begin{algorithm}[t]
\caption{\textsc{The \textsc{IFIAL} Algorithm}}
\begin{algorithmic}
\STATE \textbf{Input: }Tabular data with missing values, $\{X\in\Re^{N\times d}, y\}$
\STATE \textbf{Parameters:} $k$: number of features in partitions. 
\STATE \textbf{Output:} Trained classifier model, $\mathcal{M}$ ($\theta$) 
\vspace{0.5em}
\STATE \textbf{Procedure:}
\STATE $\{X^{obs}, X^{miss} \} \leftarrow \{ X,  y \} $
\STATE Missing rate of $f_j$ feature, $r_j \leftarrow \frac{length (X^{miss} (:, j))}{length (X(:, j))}$ 
\STATE $\{f_1, f_2,..., f_d\} \leftarrow$ $\{r_1<r_2<...<r_d\}$, sorting $f_i$.  
\STATE $\{P_1, P_2,...\}$ $\leftarrow$ $\{X_1(f_1, f_2), X_2 (f_2, f_3),...\}$, when k=2.
\vspace{0.5em}
\FOR { $P_i \in \{P_1, P_2,...\}$}
\STATE \{$X_i^{obs}, X_i^{miss}$\} $\leftarrow P_i$
\STATE $m \leftarrow X_i^{\text{miss}}$, Missing mask vector
\STATE $M_1 \leftarrow m$,~~Eq. \ref{eq:mask_m1}
\STATE $M_2 \leftarrow \bar{m}^T\bar{m}$
\STATE $\mathcal{M}(\theta) \leftarrow \{ (X_i, M_1, M_2), y_i\}$,~~Eq. \ref{eq:self-attention-final}
\ENDFOR

\end{algorithmic}
\end{algorithm}

A binary vector $m \in \Re^{1\times d}$ uses 1s to identify features with missing values and 0s otherwise. An outer product between a column vector of 1s and $m$ yields a $(d, d)$ binary matrix ($M_1$) with columns of 1s corresponding to the features with missing values. These columns of 1s are replaced with large negative values in Equation \ref{eq:mask_m1}. An exponential function transforms large negative values in $M_1$ representing missing features to zeros, as shown in Equation \ref{eq:mask_m1_exp}.
\begin{eqnarray}
\label{eq:mask_m1}
    M_1 &=& -\infty \odot (\mathbf{1}_d^T m) \\ 
    \label{eq:mask_m1_exp}
    exp (M_1)&=&
    \begin{cases}
      0, & \text{if}\ M_1(:) = -\infty \\
      1, & \text{if}\ M_1(:) = 0
    \end{cases} 
\end{eqnarray}
%
$M_1$ filters the attention columns corresponding to the features with missing values using the softmax function in Equation \ref{eq:self-attention}. The attention rows corresponding to features with missing values are eliminated using another binary mask, $M_2$. Here, $M_2 = \bar{m}^T \bar{m}$ obtained from the complement of $m$ ($\bar{m}$) includes 0s in rows and columns that correspond to missing features and 1s otherwise. 
\begin{eqnarray}
\label{eq:self-attention}
\text{Attention} &=&  \text{softmax}(\frac{Q K^T} {\sqrt{h}} + M_1) \odot M_2  \\
&=& (\frac {\exp{(\frac{Q K^T} {\sqrt{h}} + M_1})}{\sum \exp{(\frac{Q K^T} {\sqrt{h}} + M_1})} \odot M_2 )   \nonumber \\
\label{eq:self-attention-final}
&=& \frac  {\exp{(\frac{Q K^T} {\sqrt{h}}) * \exp {(M_1)}}}{\sum \exp{(\frac{Q K^T} {\sqrt{h}}) * \exp {(M_1)}}} \odot M_2
\end{eqnarray}

Combining Equations \ref{eq:mask_m1_exp} and \ref{eq:self-attention-final}, a $d\times h$ attention head in Equation 6 yields a $h$ dimensional embedding for each of the $d$ features. The query ($Q$), key ($K$), and value ($V$) matrices obtain the attention head in Equation \ref{eq:head}. 
\begin{eqnarray}\label{eq:self}
\text{\small Head =} 
\label{eq:head}
\begin{cases}
(0* I_{d\times h}) *V , &\text{\small if  $M_1(:)$\text{ = -$\infty$}, $M_2(:)=0$} \\
\text{\small softmax}(\frac{Q K^T} {\sqrt{h}}) V, &\text{\small if  $M_1(:)=0$, $M_2(:)=1$}
\end{cases}
\end{eqnarray}
$M_1$ uses $- \infty$ to mask the attention columns with missing features. The remaining rows of attention with missing features are masked by 0s of $M_2$. Otherwise, $M_1$ and $M_2$ retain the attention scores of all observed feature pairs using 0s and 1s, respectively.

\section {Experiments}
\label{experiments}
\begin{table*}[t]
\caption{Summary of tabular data sets with heterogeneous data types and structures used for benchmarking.}
\centering
\label{table_dataset}
\scalebox{0.7}{
\begin{tabular}{lllccccc}
\toprule
Type & OpenML & Data set     & Samples & Features & Numer- & Categor & Classes \\ 
     & Id     & &         &          & ical   & -ical   &         \\ 
\midrule
\multirow{13}{*}{\shortstack{Complete\\Data \\sets}} 
     & 13     & Breast-cancer    & 286     & 9        & 0         & 9           & 2    \\
     & 31     & Credit-g         & 1000    & 20       & 7         & 13          & 2    \\
     & 37     & Diabetes         & 768     & 8        & 8         & 0           & 2    \\
     & 181    & Yeast            & 1484    & 9        & 9         & 0           & 10   \\
     & 334    & Monks-problems-2 & 601     & 7        & 0         & 7           & 2    \\
     & 463    & Backache         & 180     & 31       & 5         & 26          & 2    \\
     & 1063   & Kc2              & 522     & 21       & 21        & 0           & 2    \\
     & 1067   & Kc1              & 2109    & 21       & 21        & 0           & 2    \\
     & 1071   & Mw1              & 403     & 37       & 37        & 0           & 2    \\
     & 1480   & Ilpd             & 583     & 10       & 9         & 1           & 2    \\
     & 1498   & SA-heart         & 462     & 9        & 8         & 1           & 2    \\
     & 23381  & Dresses-sales    & 500     & 12       & 1         & 11          & 2    \\
     & 40691  & Wine-quality-red & 1599    & 11       & 11        & 0           & 6    \\
\midrule
\multirow{4}{*}{\shortstack{Natural\\Missing\\Values}}
     & 55     & Hepatitis        & 155     & 19       & 6         & 13          & 2    \\
     & 6332   & Cylinder-bands   & 540     & 39       & 18        & 21          & 2    \\
     & 41162  & Kick             & 72983   & 10       & 4         & 6           & 2    \\
     & 41440  & Okcupid-stem     & 50789   & 10       & 9         & 1           & 3    \\
\bottomrule
\end{tabular}
}
\end{table*}

We perform experiments on three types of missing values: 1) missing completely at random (MCAR), 2) missing not at random (MNAR), and 3) missing values that appear naturally with the data. Here, MCAR and MNAR types are simulated at a missing rate ranging from 10\% to 50\% in increments of 10\%.

\subsection{Tabular data sets}

Table \ref{table_dataset} summarizes the tabular data sets used for the benchmarking. For generalizability, we have selected 17 diverse data sets of varying sizes and a mix of numerical and categorical features from the OpenML repository~\citep{OpenMLCC18}. Notably, tabular data sets with large sample sizes ($>$10,000) are often selected to facilitate deep learning~\citep{caruso2024, FTT_Gorishniy2021}. However, most tabular data sets in practice have relatively smaller sample sizes ($<$1000)~\citep{rabbani-g-ceals-2024}, which are more challenging data sets for deep learning methods. Therefore, our selected data sets have sample sizes ranging from 155 to 72983, while the feature dimension ranges from 7 to 39. The first 13 data sets are available in their complete form, without missing values, and are used to simulate the MCAR and MNAR types of missing values following~\citep{samad2022}. The other four data sets with IDs 55, 6332, 41162 and 41440 have natural missing values at the rate of 48.8\%, 48.7\%, 30.2\%, and 20.6\%, respectively.

\subsection{Model implementation and evaluation}

The FTT classifier, adopted from~\citep{Wang2024Transtab}, consists of a transformer with two encoder layers. The encoder layers transform numerical and categorical features into a unified 128-dimensional embedding. Each Transformer layer includes eight attention heads and a feedforward network with a 2048-dimensional hidden layer, ReLU activation, and a dropout rate of 0.3. We have used the Adam optimizer with a fixed learning rate of \(1 \times 10^{-5}\) and no weight decay. The model is trained for up to 300 epochs using a batch size of 128. Early stopping is used with a patience of 50 epochs, terminating training when the validation loss stops improving for 50 consecutive epochs. The proposed and all baseline models are validated using Optuna~\citep{optuna_akiba}. For each experiment, we run 100 Optuna trials, randomly sampling different hyperparameter settings and evaluating them on the validation split. The model achieving the best validation performance is then selected and evaluated on the corresponding test fold. The hyperparameter search spaces are reported in Table~\ref{tab:hyperparameters}.

A five-fold cross-validation yields the average area under the receiver operating characteristic (ROC) curve (AUC). The average rank order of each method is reported in 13 tabular data sets based on AUC scores~\citep{Borisov2022}.  The five missing rates from 10\% to 50\% in 13 data sets yield 65 experimental scenarios for each missing value type. As in~\citep{tokar2024icet}, we use a win matrix to compare methods in pairs, showing the percentage of experimental scenarios in which one method outperforms the other.  The resilience of a model to varying missing rates is evaluated by comparing the AUC scores against the reference AUC score. The reference AUC score of a data set is obtained in its complete form before simulating missing values. We primarily evaluate the proposed incremental learning method for three partition sizes, k = 2, 3, and 4. Our analysis in Figure~\ref{fig:computations-comparison} suggests that FIL using feature partitions is computationally more efficient than learning the complete feature space ($d$ features) when k is less than $\lceil d/2 \rceil$. Therefore, we report the performance of our model for partitions with $k = \frac{d}{2}$.

\subsection {Baseline methods}
Our baseline methods can be categorized into two types: one requires imputation before classification, and the other can classify data directly without a standalone imputation. The imputation requirement before classification is satisfied by median value imputation, MICE~\citep{Lauren2021}, GAN-based imputation method GAIN~\citep{yoon2018gain}, and diffusion-based imputation method, Diffputer~\citep{diffputer}. Unlike mean values, median values are robust to outliers and are recommended for low missing rates. The classification of imputed data sets is achieved using two state-of-the-art models for tabular data, Gradient boosting trees (GBTs) and FTT~\citep{Wang2024Transtab}. On the other hand, XGBoost~\citep{chen2016xgboost} and LightGBM~\citep{ke2017lightgbm} can use the Missing Incorporated in Attributes (MIA) strategy~\citep{twala2008mia}, as mentioned in Section~\ref{related_work}, to directly classify data with missing values. XGBoost and LightGBM treat missing values as distinct categories during decision tree splits. The FTT classifier can directly classify missing values using an attention mask, presented in Equation~\ref{eq:self-attention}.  Therefore, an AM-FTT framework, similar to ~\citep{caruso2024}, is considered a baseline. The proposed IFIAL method is compared against 11 baseline methods: 1) median - GBT, 2) median - FTT, 3) MICE - GBT, 4) MICE - FTT, 5) GAIN - GBT, 6) GAIN - FTT, 7) Diffputer - GBT, 8) Diffputer - FTT, 9) AM - FTT, 10) MIA - XGBoost, and 11) MIA - LightGBM.

\section{Results}
\label{results}
All results are obtained using an Ubuntu 22.04 machine utilizing an Intel(R) Xeon(R) W-2265 CPU (24 logical cores) running at 3.70GHz, 64GB of RAM, and a Quadro RTX A4000 GPU with 16GB of video memory. The performance of the proposed IFIAL method is compared with the baseline methods using 1) the average rank based on the AUC scores, 2) the win matrix for pairwise comparisons of the methods, and 3) the robustness of the model to higher missing rates.~\ref{AUC_F1} presents the average AUC scores for individual data sets with the MCAR type in Table~\ref{auc-mcar} and the MNAR type in Table~\ref{auc-mnar}.

\begin{table*}[t]
\caption{Average performance rank of the proposed (IFIAL) and baseline methods across 13 diverse MCAR-type data sets with five missing value rates.}
\label{rank-table-mcar}
\centering
\scalebox{0.65}{
\begin{tabular}{lcccccccc}
\toprule
Method        & Imputation & 10\%      & 20\%       & 30\%       & 40\%       & 50\%       & Avg. Rank  & Overall Rank \\ 
\midrule
Median-GBT     & Yes        & 9.0(2.62)  & 7.0(4.17)  & 7.9(5.33) & 7.2(4.46) & 7.1(5.03)  & 7.7(0.74) & 9            \\
Median-FTT     & Yes        & 7.6(4.82)  & 7.9(4.42)  & 8.2(4.74) & 7.8(5.41) & 8.8(4.40)  & 8.1(0.42) & 10           \\
MICE-GBT       & Yes        & 7.4(3.54)  & 7.6(4.84)  & 7.5(4.47) & 8.0(4.24) & 7.0(4.41)  & 7.5(0.33) & 8            \\
MICE-FTT       & Yes        & 6.8(2.20)  & 6.6(2.37)  & 5.6(3.69) & 6.4(3.05) & 6.0(3.37)  & 6.3(0.43) & 4            \\
GAIN-GBT       & Yes        & 8.4(5.55)  & 10.6(3.44) & 6.8(4.02) & 8.4(5.13) & 7.8(4.55)  & 8.4(1.25) & 13           \\
GAIN-FTT       & Yes        & 6.7(4.15)  & 6.0(4.28)  & 6.0(4.86) & 6.9(4.67) & 6.6(2.64)  & 6.4(0.36) & 5            \\
Diffputer-GBT  & Yes        & 9.5(3.96)  & 8.1(4.52)  & 8.5(4.90) & 7.0(4.93) & 7.4(4.90)  & 8.1(0.88) & 11           \\
Diffputer-FTT  & Yes        & 7.4(2.84)  & 8.6(2.06)  & 9.2(2.45) & 9.1(1.88) & 8.4(2.59)  & 8.5(0.66) & 14           \\
\midrule
MIA-Xgboost    & No         & 8.4(4.51)  & 7.6(5.18)  & 7.8(3.96) & 6.4(5.94) & 6.4(5.90)  & 7.3(0.80) & 7            \\
MIA-LightGBM   & No         & 10.0(2.83) & 9.0(2.83)  & 8.5(4.95) & 9.0(5.66) & 10.0(4.24) & 9.3(0.60) & 15           \\
AM-FTT         & No         & 7.1(2.67)  & 8.3(3.90)  & 8.7(2.50) & 7.9(3.18) & 9.6(4.50)  & 8.3(0.82) & 12           \\
IFIAL (K =2)   & No         & 4.9(4.29)  & 4.8(2.43)  & 5.5(2.51) & 5.4(2.83) & 6.2(2.71)  & 5.3(0.53) & 2            \\
IFIAL (K =3)   & No         & 5.5(5.07)  & 6.2(3.65)  & 5.9(3.00) & 6.9(3.56) & 6.4(3.34)  & 6.2(0.47) & 3            \\
IFIAL (K =4)   & No         & 6.8(3.96)  & 6.9(4.58)  & 7.0(4.04) & 7.4(3.62) & 6.5(3.74)  & 6.9(0.29) & 6            \\
IFIAL (K =d/2) & No         & 4.0(3.66)  & 4.1(3.60)  & 4.6(3.38) & 4.5(2.62) & 4.8(3.49)  & 4.4(0.29) & 1            \\ 
\bottomrule
\end{tabular}}
\end{table*}

\begin{table*}[htbp]
\caption{Average performance rank of the proposed (IFIAL) and baseline models across 13 diverse MNAR-type data sets with five missing value rates.}
\label{rank-table-mnar}
\centering
\scalebox{0.65}{
\begin{tabular}{lcccccccc}
\toprule
Method        & Imputation & 10\%      & 20\%       & 30\%       & 40\%       & 50\%       & Avg. Rank  & Overall Rank\\ 
\midrule
Median-GBT     & Yes        & 7.1(3.73)  & 5.9(3.93)  & 6.7(4.74)  & 5.5(4.25)  & 5.1(4.86)  & 6.1(0.74)  & 7            \\
Median-FTT     & Yes        & 6.2(3.94)  & 6.4(4.03)  & 6.4(4.20)  & 6.3(4.60)  & 6.5(4.02)  & 6.4(0.10)  & 10           \\
MICE-GBT       & Yes        & 6.5(3.46)  & 7.1(4.52)  & 7.5(3.82)  & 7.6(4.00)  & 6.4(4.21)  & 7.0(0.51)  & 11           \\
MICE-FTT       & Yes        & 5.4(3.62)  & 5.4(3.50)  & 4.6(3.32)  & 5.0(3.35)  & 4.6(3.43)  & 5.0(0.39)  & 3            \\
GAIN-GBT       & Yes        & 8.4(5.55)  & 10.8(3.63) & 7.2(4.82)  & 8.8(5.07)  & 7.6(4.56)  & 8.6(1.25)  & 14           \\
GAIN-FTT       & Yes        & 4.9(3.62)  & 4.3(3.28)  & 4.7(3.91)  & 5.2(3.77)  & 5.0(3.35)  & 4.8(0.30)  & 2            \\
Diffputer-GBT  & Yes        & 8.8(4.33)  & 7.8(4.20)  & 8.5(4.34)  & 6.8(4.40)  & 6.8(4.68)  & 7.7(0.84)  & 12           \\
Diffputer-FTT  & Yes        & 5.6(3.60)  & 6.5(3.43)  & 6.5(3.72)  & 6.6(3.60)  & 5.9(3.61)  & 6.2(0.39)  & 8            \\
\midrule
MIA-Xgboost    & No         & 8.6(4.83)  & 8.0(4.80)  & 8.6(3.36)  & 7.2(5.81)  & 6.8(5.85)  & 7.8(0.73)  & 13           \\
MIA-LightGBM   & No         & 11.5(0.71) & 10.0(1.41) & 10.5(2.12) & 11.0(2.83) & 12.0(1.41) & 11.0(0.71) & 15           \\
AM-FTT         & No         & 5.9(3.33)  & 6.7(3.54)  & 6.3(2.92)  & 6.0(3.08)  & 6.9(3.92)  & 6.4(0.38)  & 9            \\
IFIAL (K =2)   & No         & 5.4(2.61)  & 4.8(2.68)  & 5.4(2.88)  & 7.0(0.82)  & 7.0(0.82)  & 5.9(0.91)  & 5            \\
IFIAL (K =3)   & No         & 3.8(1.79)  & 6.0(2.00)  & 4.4(2.30)  & 5.2(2.59)  & 6.0(1.87)  & 5.1(0.87)  & 4            \\
IFIAL (K =4)   & No         & 5.8(3.56)  & 5.2(2.86)  & 6.2(2.59)  & 5.8(2.68)  & 7.2(2.95)  & 6.0(0.66)  & 6            \\
IFIAL (K =d/2) & No         & 3.3(2.66)  & 2.0(1.55)  & 2.8(2.64)  & 3.7(2.66)  & 5.5(2.43)  & 3.5(1.16)  & 1            \\
\bottomrule
\end{tabular}}
\end{table*}

\begin{table*}[t]
\caption{Average AUC scores for four data sets with natural missing values: (Hepatitis (ID: 55), Cylinder-bands (ID: 6332), Kick (ID: 41162), Okcupid-stem (ID: 41440)) with rates of 48.8\%, 48.7\%, 30.2\%, and 20.6\%, respectively.}
\label{rank-table-natural-missing}
\scalebox{0.45}{
\begin{tabular}{llllllllllllll}
\toprule
ID    & Median- & Median- & MICE- & MICE- & GAIN- & GAIN- & Diffputer- & Diffputer- & MIA- & MIA- & AM- & IFIAL \\
      & GBT & FTT & GBT & FTT & GBT & FTT & GBT & FTT & Xgboost & LightGBM & FTT & (K=d/2) \\
\midrule
55    & 0.815 (0.08) & 0.809 (0.04) & 0.803 (0.09) & 0.842 (0.04) 
      & 0.792 (0.11) & 0.839 (0.05) & 0.801 (0.10) & 0.808 (0.06)
      & 0.765 (0.06) & 0.790 (0.07) & 0.808 (0.04) & 0.848 (0.04) \\
6332  & 0.872 (0.02) & 0.748 (0.01) & 0.874 (0.02) & 0.801 (0.04)
      & 0.751 (0.07) & 0.812 (0.03) & 0.846 (0.03) & 0.786 (0.05)
      & 0.918 (0.01) & 0.765 (0.06) & 0.818 (0.03) & 0.868 (0.03) \\
41162 & 0.744 (0.02) & 0.739 (0.01) & 0.744 (0.01) & 0.742 (0.01)
      & 0.740 (0.02) & 0.741 (0.03) & 0.746 (0.01) & 0.741 (0.01)
      & 0.741 (0.02) & 0.745 (0.01) & 0.742 (0.01) & 0.747 (0.01) \\
41440 & 0.689 (0.01) & 0.659 (0.02) & 0.692 (0.01) & 0.676 (0.01)
      & 0.656 (0.03) & 0.654 (0.02) & 0.687 (0.01) & 0.659 (0.03)
      & 0.670 (0.01) & 0.687 (0.01) & 0.675 (0.01) & 0.681 (0.01) \\
\midrule
\multicolumn{1}{c}{\begin{tabular}{@{}c@{}}Avg.\\Rank\end{tabular}} 
  & 4.0 (0.03) & 7.3 (0.02) & 4.0 (0.04) & 5.0 (0.04)
  & 9.0 (0.05) & 5.2 (0.02) & 6.5 (0.04) & 6.8 (0.03)
  & 7.8 (0.02) & 5.8 (0.04) & 5.0 (0.02) & 2.3 (0.02) \\
\bottomrule
\end{tabular}}
\end{table*}

\subsection {Average performance rank}

Table \ref{rank-table-mcar} compares the proposed IFIAL method with 11 baselines for MCAR-type data. In single-value imputation, such as median value imputation, the GBT classifier performs better than FTT, regardless of missing rates. The MICE (multiple imputation) method outperforms median-value imputation. The GAN-based imputation approach (GAIN) using the FTT classifier produces better performance than MICE at low levels of missingness (10\% and 20\%), but its performance deteriorates relative to MICE as the missing rate increases. In comparison, the diffusion-based imputation approach performs worse overall, with both Diffputer variants consistently appearing among the lower-performing methods across all missing rates. Among the methods without standalone imputations, the MIA-based model (MIA-XGBoost) outperforms AM-FTT, confirming the known superiority of traditional ML over DL. The average performance rank shows that the proposed IFIAL (K = d/2) outperforms all 11 baselines. IFIAL with other feature partition sizes, including IFIAL (K = 2) and IFIAL (K = 3), also rank among the top three. Even at low missing rates (10\%–30\%), IFIAL (K = d/2) maintains a strong performance rank, while at higher rates it shows clear superiority. These results confirm the robustness and consistent performance of the proposed IFIAL method for MCAR-type data. Table~\ref{rank-table-mnar} reports the mean classification performance ranks for models applied to MNAR-type data. Imputation-based approaches that already perform well on MCAR data (GAIN-FTT and MICE-FTT) also achieve strong results on MNAR data, where they rank 2nd and 3rd, respectively. In contrast, the imputation-free MIA models (MIA-Xgboost and MIA-LightGBM) perform worse on MNAR data than on MCAR data. This observation suggests that while MIA-based models may benefit from informative missingness (MNAR) in large-sample size scenarios~\citep{perez2022_imputationbenchmarking}, their performance can degrade when applied to relatively small data~\citep{le2025_imputation}. However, for MNAR-type data, Median-GBT consistently outperforms MIA-based methods across all missing rates, unlike in MCAR-type scenarios. In contrast, the proposed IFIAL with several partition sizes (K = 2, 3) ranks among the top five methods, while the partition (K = d / 2) produces the best performance, as it does for the MCAR-type data. These results suggest that the proposed IFIAL method is not substantially affected by missing value types, unlike other imputation-based approaches. Table~\ref{rank-table-natural-missing} presents the results for the four data sets with naturally occurring missing values. In this setting, the proposed IFIAL method (with K = d/2) attains the best average performance rank (2.3 (0.02)), outperforming the second-best method, median imputed GBT classification (4.0 (0.04)).  Interestingly, MIA-Xgboost shows the best on a data set (ID 6332) with limited samples but a relatively large feature space. However, for data sets with large sample sizes, differences in model performance are insignificant. The availability of a large number of samples provides sufficient information and eliminates the challenge of learning data sets with small sample sizes.

\begin{figure*}[t]
\centering
\subfigure[MCAR type] { \includegraphics[trim=1.0 0 0 0cm, clip, width=0.5\textwidth] {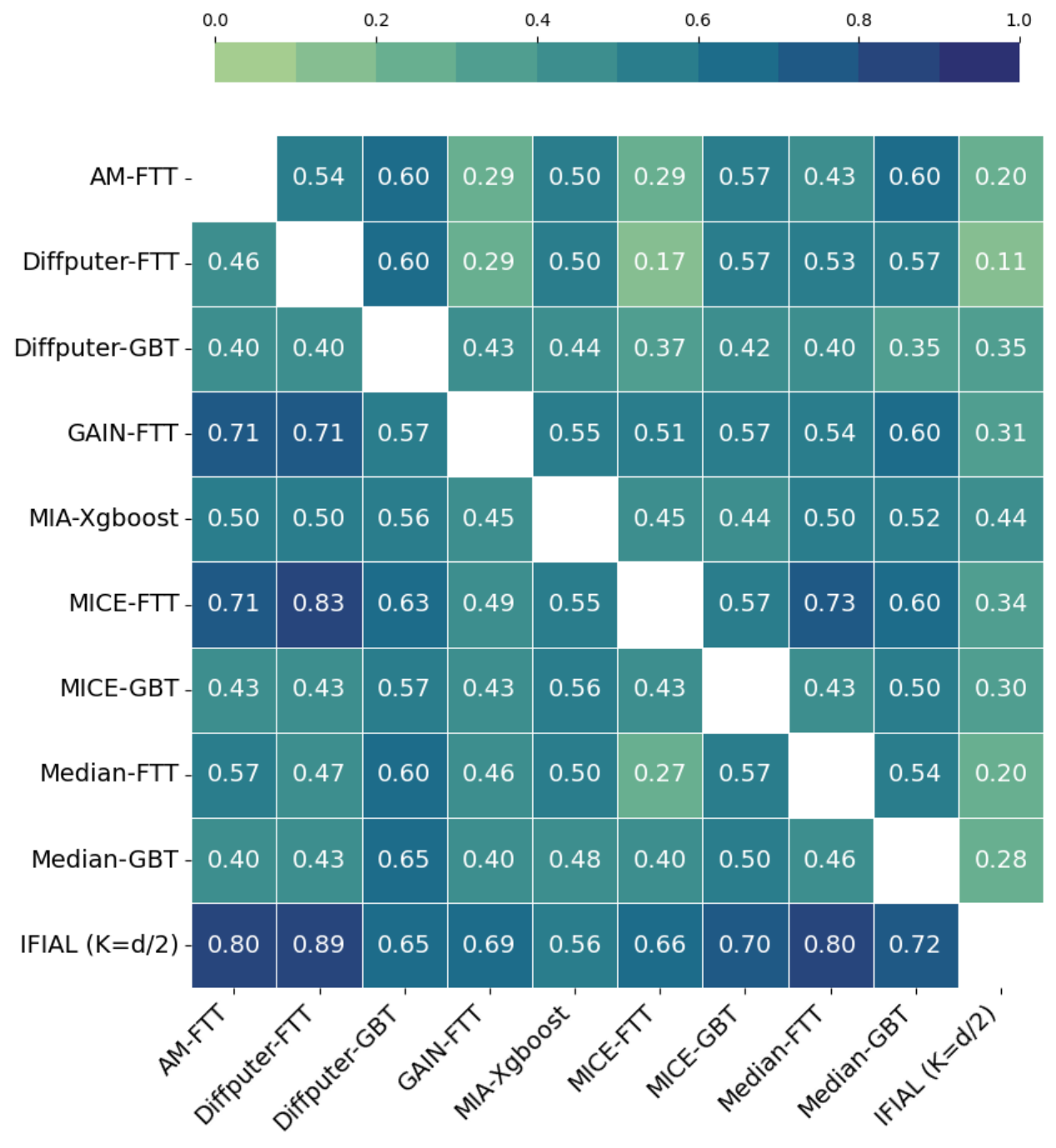}}
\hspace{-15pt}
\subfigure[MNAR type] { \includegraphics[trim=0.0 0 0 0cm, clip, width=0.5\textwidth] {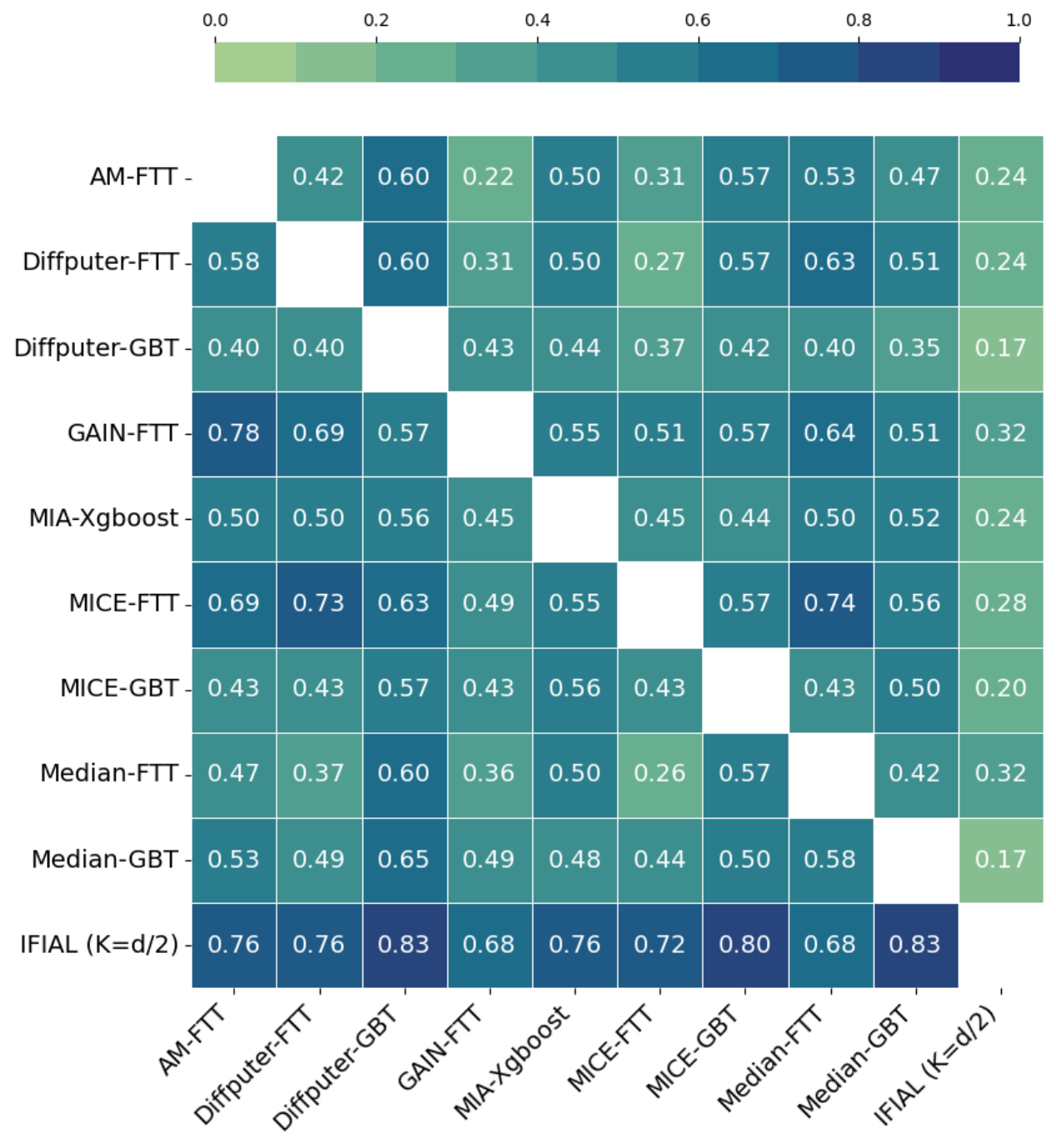}}
\caption{Win matrix. Values are the fraction of experimental scenarios in which the row methods outperform the methods in the columns in terms of AUC scores.}
\label{win_matrix}
\end{figure*}

\subsection {Win matrix, computational time, and robustness}

We count the experimental scenarios in which one method wins over another. There are 65 experimental scenarios for each missing value type using 13 data sets and five missing rates. Figure~\ref{win_matrix} (a) for the MCAR-type data shows that the proposed IFIAL method is superior to the baselines for more than 50\% of the experimental scenarios. In the most favorable setting, IFIAL surpasses Diffputer-FTT in 89\% of scenarios. Figure~\ref{win_matrix} (b), which presents the results for MNAR-type data, also shows that IFIAL surpasses all baseline methods in more than 50\% of experimental scenarios. Furthermore, IFIAL outperforms Diffputer-GBT and Median-GBT in 83\% of the scenarios. IFIAL also outperforms MIA-Xgboost in 76\% of scenarios, which is 56\% for MCAR-type data. These comparisons indicate that deep-imputation methods are not optimized for MNAR-type data.

\subsection {Computational time and robustness to missing rates}

\begin{figure*}[t]
\centering
\subfigure[MCAR type\label{resilience-plot_MCAR}] { \includegraphics[trim=0 0 0 0cm, clip, width=0.48\textwidth] {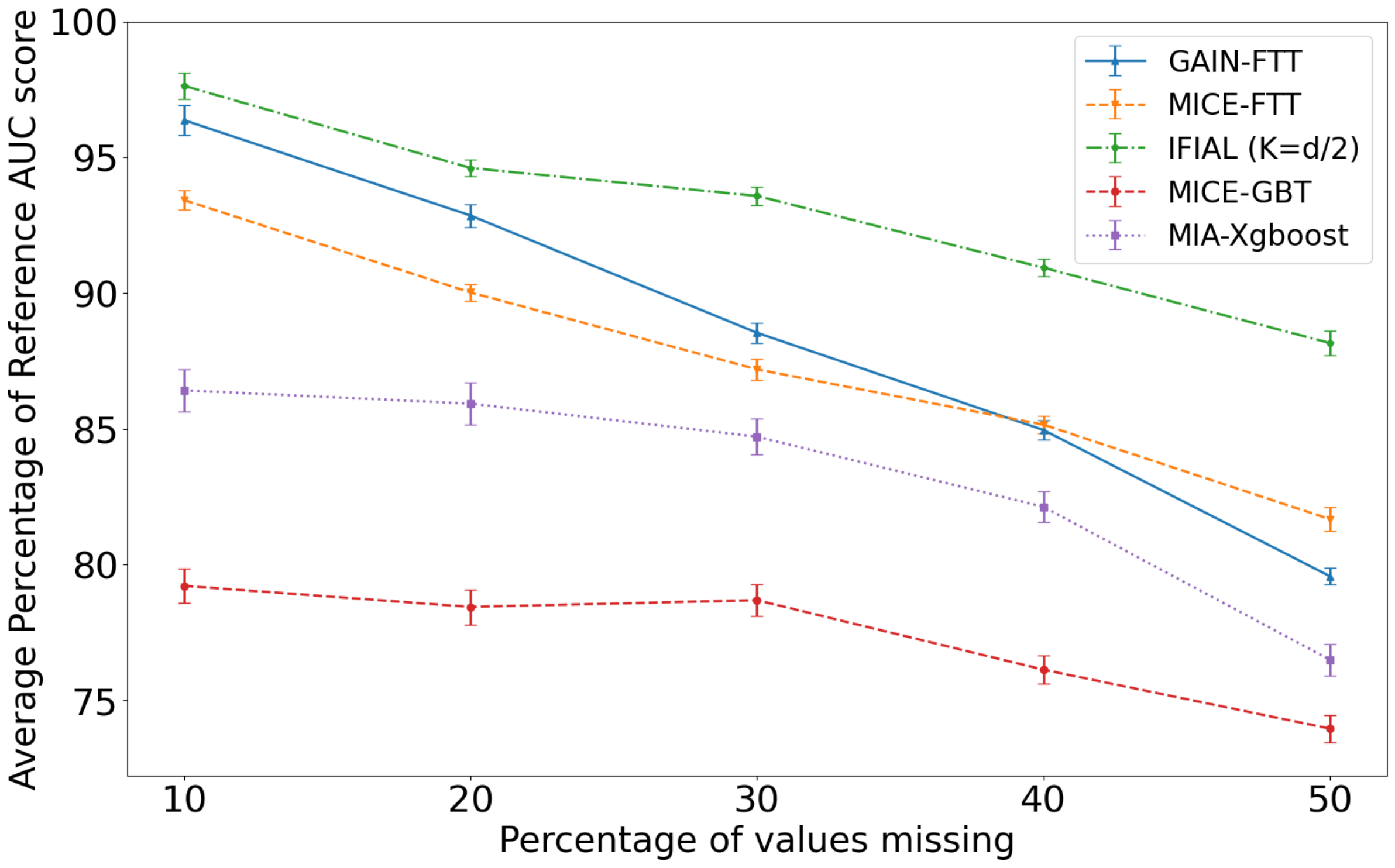}}
\hspace{0pt}
\subfigure[MNAR type\label{resilience-plot_MNAR}] { \includegraphics[trim=0 0 0 0cm, clip, width=0.48\textwidth] {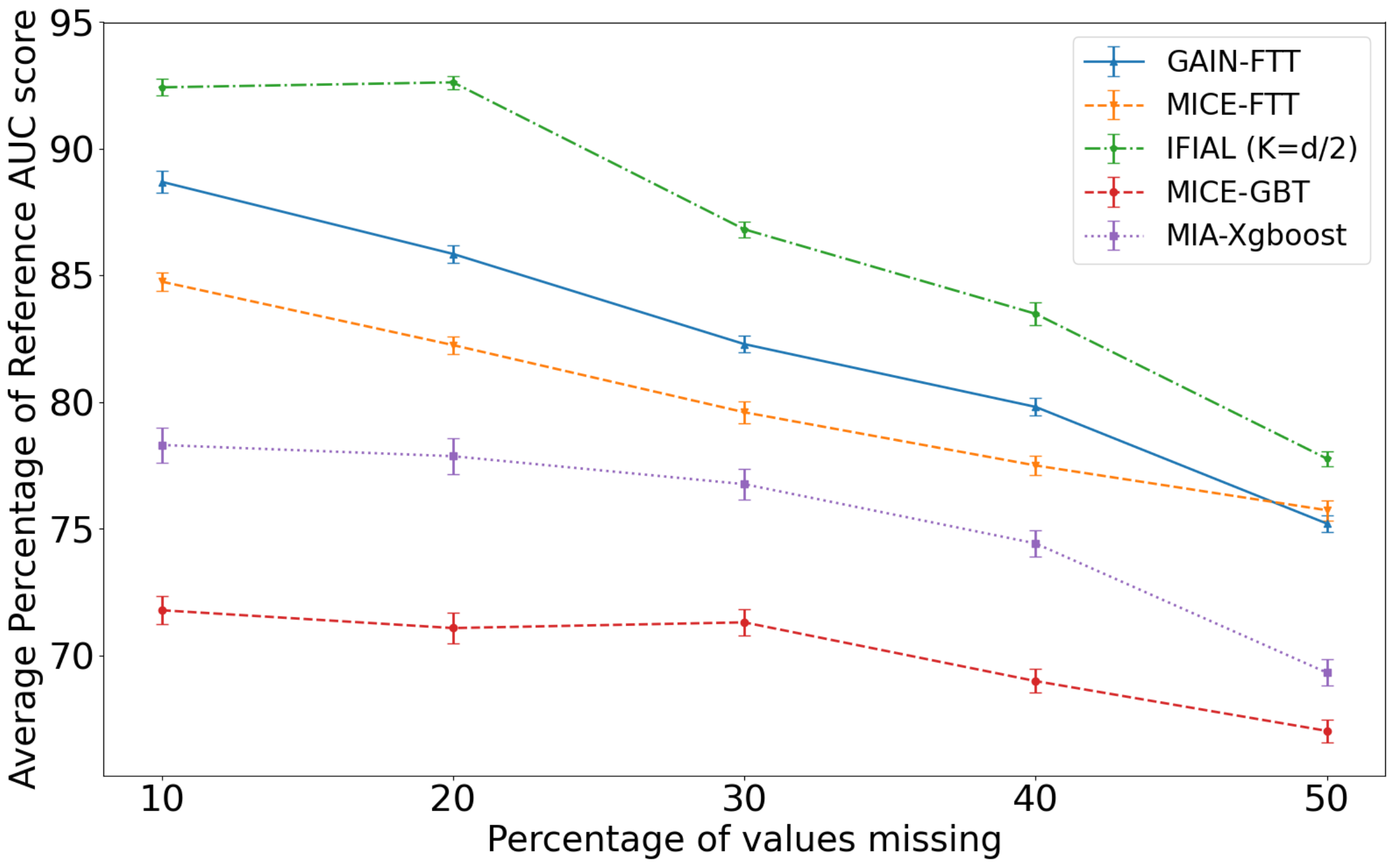}}
\vspace{-10pt}
\caption{Effect of increasing missing value rates. The average percentage of the reference AUC score is obtained across 13 data sets.}
\label{resilience-plot}
\end{figure*}

Table~\ref{computational} presents the run time for the imputation and classification steps of the baseline methods. As expected, methods faster than IFIAL include GAIN-GBT, Diffputer-GBT, AM-FTT, MIA-Xgboost, and MIA-LightGBM, mainly due to the fast GBT classifier model. However, the classification performance rank of these methods is worse than that of IFIAL. One computational benefit of IFIAL is that it saves the time required for imputation. Figure~\ref{resilience-plot} shows the effect of varying the missing rates on the performance of five top performing methods.  The baseline AUC values used in the robustness analysis are derived from classifications performed on full data sets with 0\% missing values. The AUC score decreases relative to the reference AUC as the missing rate increases. Figure~\ref{resilience-plot_MCAR} for MCAR-type data shows better robustness of the proposed method when the missing rate is greater than 30\%.  However, Figure~\ref{resilience-plot_MNAR} for MNAR-type data demonstrates that the proposed IFIAL method preserves its robustness even as the missing rate increases, clearly outperforming all other competitive baselines.

\begin{table*}[!t]
\vspace{0pt }
\centering
\caption{Computational runtime in seconds for imputation (training and imputing) and classification (training and inference) steps using the credit-g data set with 50\% MCAR type missing values. 
}
\label{computational}
\scalebox{0.7}{
\begin{tabular}{lccccc}
\toprule
Method & Imputation & Classification & Total & Classification & Classification  \\
& runtime & runtime & runtime & Rank (MCAR) & Rank (MNAR)\\
\midrule
MICE - GBT & 417 & 2 & 419& 8& 11\\
MICE - FTT & 417 & 190 & 607 & 4 & 3\\
GAIN - GBT & 51 & 2 & 54 & 13&14\\
GAIN - FTT & 51 & 190 & 241 & 5 &2\\
Diffputer - GBT & 8 & 1 & 9 & 11&12\\
Diffputer - FTT & 8 & 190 & 198 & 14&8\\
AM-FTT & N/A & 21 & 21  & 12& 9\\
MIA - Xgboost & N/A & 1 & 1&7 &13\\
MIA - LightGBM & N/A & 1 & 1& 15 & 15\\
IFIAL (k = $\frac{d}{2}$) & N/A & 163 & 163 & 1&1\\
\bottomrule
\end{tabular}}
\end{table*}

\subsection{Ablation studies}
\begin{figure*}[t]
    \centering
    \includegraphics[width=1.0\linewidth]{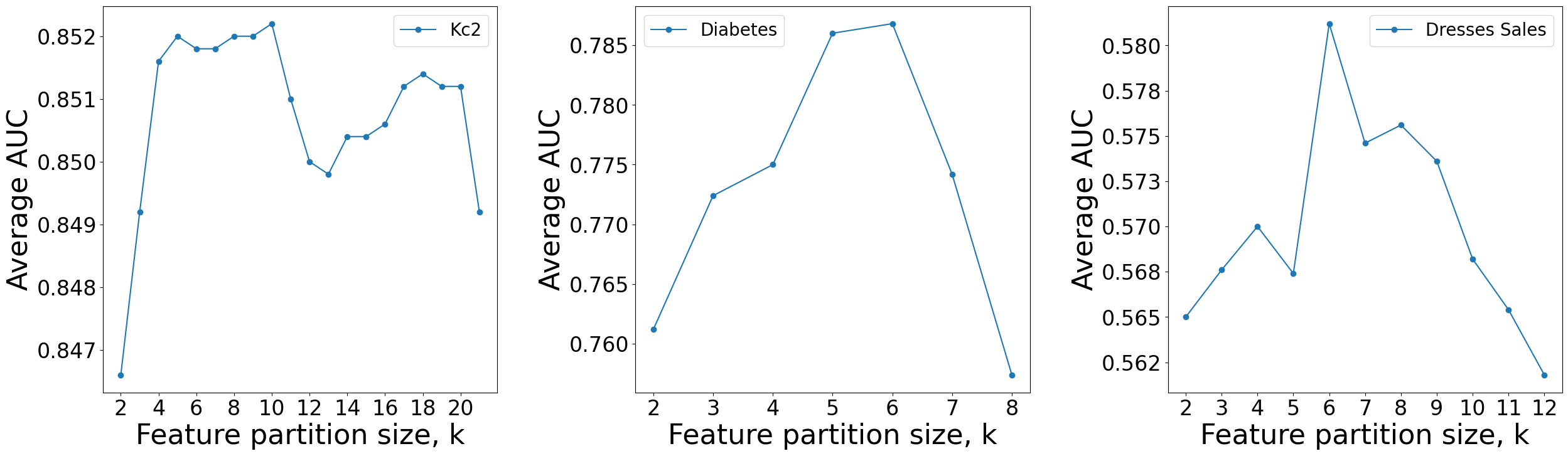}
    \caption{Effects of partition size (k) on average AUC scores obtained across varying missing value rates. The total number of features (\( d \)) for three data sets are: the Kc2 data set (\( d = 21 \)), Diabetes data set (\( d = 8 \)), and Dresses Sales data set (\( d = 12 \)).}
    \label{fig:auc_vs_k}
\end{figure*}

In an ablation study, we vary the feature partition size (k) for three representative data sets: Kc2 (ID 1063), Diabetes (ID 37), and Dresses sales (ID 23381). Figure~\ref{fig:auc_vs_k} shows the k values that yield the best AUC scores for individual data sets. In general, incremental learning of features is superior to learning with all features at once (that is, without incremental learning, k = d). The best average AUCs are obtained for the data sets Kc2 $(d = 21)$ at k=10, Diabetes $(d = 8)$ at k=6, and Dresses sales $(d = 12)$  at k = 6. Therefore, the value of k can be chosen as half of the feature dimension for the best classification performance. This observation aligns with the computational cost analysis presented in Figure~\ref{fig:computations-comparison} in ~\ref{App_A}, where $k < \frac{d}{2}$ is computationally more efficient than using the entire feature space ($d$). In general, IFIAL outperforms all or most of the baselines at k = 3, which can be further improved by increasing the size of the partition to $\frac{d}{2}$.

\section {Discussions}
\label{discussions}

The paper proposes incremental learning for tabular data to eliminate the requirement of initializing, imputing, and completing missing values. The findings of the paper can be summarized as follows. First, the proposed IFIAL method outperforms all baseline methods with and without data imputation, regardless of the missing value types. Second, state-of-the-art imputation methods are sensitive to missing value types, especially when missing values are not at random (MNAR).  The proposed IFIAL method entirely avoids the imputation of missing values and, therefore, is a superior choice when the missing value type is unknown or complex. Third, while retrofitting a pair of attention masks enables imputation-free learning of tabular data representations, incremental learning of feature partitions is essential for overall efficiency and performance. In comparison, the performance of AM-FTT suggests that attention masks and the FTT model are insufficient to achieve optimal performance. Fourth, the IFIAL method offers the best performance and computational time trade-off with better resilience to missing value rates than baselines. In general, our proposed method renders existing missing value imputation methods unnecessary from both computational efficiency and performance standpoints.

\subsection{Robustness to missing value types}
The proposed method brings several advantages over existing imputation methods. First, imputation methods are usually not selected and designed based on different missing value types. Therefore, an effective imputation method for MCAR-type data may not be optimal for MNAR-type data or data that have natural missing values. The imputation error and variance in the data due to varying missing value types can be avoided using our imputation-free learning method. Furthermore, IFIAL maximizes the use of observed values in learning, thereby preserving data integrity. This characteristic is particularly valuable in sensitive domains such as healthcare and scientific research, where model-generated synthetic values are often considered unreliable and can compromise the credibility of data-driven outcomes. Second, imputation of heterogeneous feature space in tabular data (e.g. numerical, categorical) requires different learning frameworks. For example, continuous numerical value imputation is a regression problem, whereas categorical variables need classification models. The proposed method eliminates the need for separate computational steps and the requirement of train-test splits to build and evaluate the imputation model.

\subsection{Robustness to high missing rates}
Imputation-based methods typically learn missing values from their relationship with observed values. However, as the missing value rate increases, the relationship ceases to exist, causing a decline in imputation quality and downstream predictive accuracy. Many existing studies evaluate imputation methods up to a 30\% missing rate~\citep{lee2023, aimnet}. The robustness of many similar methods to higher missing rates remains unknown. Real-world scenarios may include important features missing in more than 50\% of the samples, as observed in~\citep{Samad2019Jaac}. Imputation methods are prone to higher error rates, leading to more unreliable data for downstream classification. However, methods without requiring imputation, such as AM-FTT, rely solely on observed values. The proposed IFIAL approach skips the imputation step and instead leverages feature attention via the FTT framework and incremental learning of observed values to remain robust against increasing missing value rates.

\subsection{Comparison with the literature}

The literature explores a variety of DL/ML methods for missing-value imputation. A recent survey found that deep generative models (e.g., GAIN, VAE) require a large sample size ($n > 30,000$); otherwise, traditional imputation methods (e.g., MICE) tend to outperform generative approaches~\citep{sun2023deep}. Similar sample size requirements for imputation methods do not apply to IFIAL. This may explain why MICE is superior to GAIN in our results for tabular data sets with a decent sample size. Furthermore, the recent diffusion-based method, Diffputer~\citep{diffputer} provides a comprehensive benchmarking against multiple advanced baselines, including diffusion-based methods (MissDiff~\citep{ouyang2023missdiff}, TabCSDI~\citep{tabcsdi}), generative adversarial methods (GAIN~\citep{yoon2018gain}, MIWAE~\citep{mattei2019miwae}), iterative imputation (MICE\citep{mice}, EM, HyperImpute~\citep{hyperimpute}), and graph neural network-based approaches (MIRACLE~\citep{kyono2021miracle}, GRAPE~\citep{grape}). Diffputer demonstrates better performance than these baselines, particularly highlighting its suitability for large data sets ($n > 10,000$). Our results suggest that Diffputer performance may not generalize well across different data sets, missing value rates, and types, unlike the proposed IFIAL method. It may be inferred that IFIAL would also outperform advanced imputation baselines used in benchmarking the Diffputer performance. Despite the advantages of deep imputation methods in large-scale datasets, practitioners must carefully consider the missing value rates, types, and data characteristics when selecting appropriate imputation methods for real-world applications. In particular, the choice of the imputation technique affects the quality of the data and can produce varying classification results for the same dataset and classifier pair. For example, in data set ID 463 with 50\% missingness of MCAR type, the FTT AUC score decreases markedly from 0.670 to 0.571 when median imputation is applied instead of MICE. In contrast, for data set ID 31 with 10\% missingness of MNAR type, the AUC score drops from 0.745 to 0.687 when MICE imputation is used rather than the median. The complete removal of an imputation step, as suggested by IFIAL, preserves data quality while still delivering strong and competitive classification performance.

\subsection{Broader impact}
This paper introduces a novel machine learning method for tabular data with missing values. Missing values appear in varying rates and types in many application domains, where one imputation method may not be optimal for all scenarios. Missing values in tabular data sets critically challenge the feasibility and performance of machine learning models. The widespread adoption of arbitrary missing value imputation strategies raises many questions about data quality and information reliability among data stakeholders.  Our proposed ML method enables machine learning without missing value imputation, alleviating data wastage, biases, errors, and computational costs associated with missing value imputations. The proposed method is expected to have positive social impacts through greater inclusivity, integrity, trustworthiness, and efficiency in data analysis workflows. The impact on health science and the analysis of health records data is significant, where missing values are a major barrier to the development of data-driven precision medicine. However, practitioners should consider ethical implications, such as the risks of reinforcing biases in the original data set or overlooking the reasons behind missing data.

\subsection{Limitations}
The proposed IFIAL method performs remarkably well in deep learning with tabular data, eliminating the need to impute and initialize missing values. Although deep learning without imputation of missing values is a significant contribution, the proposed method has several limitations. When the missing value rate is low and completely random, sophisticated methods for imputing missing values can yield better results than our proposed imputation-free method. Several imputation and classification models are computationally more efficient than the proposed non-imputation classification method. The performance gain of the proposed method over other methods diminishes when the sample size is very large, particularly at a low missing rate.

\section {Conclusion}
\label{conclusions}
This paper introduces IFIAL, a novel imputation-free deep learning algorithm that is performance-wise effective, robust to missing value rates and types, and computationally efficient for learning tabular data with missing values. Although attention masks in the feature tokenized transformer enable imputation-free learning of tabular data with missing values, incremental learning in feature partitions additionally yields better performance and computational efficiency. The proposed IFIAL algorithm outperforms state-of-the-art deep imputation methods in downstream classification tasks, in addition to its ability to learn the high-dimensional feature space in small feature partitions. Empirically, we demonstrate that the most computationally efficient and performance-wise adequate feature partition size for incremental learning is half the dimension of the feature space of the data set. Future work should aim at investigating imputation-free solutions for other deep learning methods.

\section{Acknowledgments}

\newpage

\bibliography{imputation}

\bibliographystyle{elsarticle-num}


\newpage
\appendix
\onecolumn

\section{Optuna Configurations}

\begin{table}[H]
\centering
\caption{Hyperparameter search space used in Optuna for IFIAL, FTT and tree-based baselines.}
\label{tab:hyperparameters}
\scalebox{0.65}{
\begin{tabular}{ll}
\toprule
\textbf{Parameter} & \textbf{Distribution} \\
\midrule

\multicolumn{2}{l}{\textit{IFIAL and FTT}} \\ \midrule
\quad num\_attention\_head & Categorical\{4, 8\} \\
\quad hidden\_dim & Categorical\{64, 128, 256\} \\
\quad num\_layer & UniformInt[2, 6] \\
\quad ffn\_ratio & Categorical\{2, 4\} \\
\quad activation & Categorical\{"relu", "gelu", "leakyrelu"\} \\
\quad hidden\_dropout\_prob & UniformFloat[0.0, 0.5] (step=0.1) \\
\quad weight\_decay & LogUniform[1e$^{-6}$, 1e$^{-2}$] \\
\quad learning\_rate & LogUniform[1e$^{-5}$, 1e$^{-3}$] \\
\quad batch\_size & Categorical\{32, 64, 128, 256\} \\
\quad warmup\_ratio & Categorical\{0.0, 0.05, 0.1\} \\
\midrule

\multicolumn{2}{l}{\textit{Gradient Boosting Trees (GBT)}} \\ \midrule
\quad n\_estimators        & UniformInt[50, 10000] \\
\quad learning\_rate       & LogUniform[1e$^{-3}$, 0.3] \\
\quad max\_depth           & UniformInt[1, 10] \\
\quad min\_samples\_split  & UniformInt[2, 20] \\
\quad min\_samples\_leaf   & UniformInt[1, 20] \\
\quad subsample            & UniformFloat[0.2, 1.0] \\
\quad max\_features        & UniformFloat[0.2, 1.0] \\
\midrule

\multicolumn{2}{l}{\textit{XGBoost}} \\ \midrule
\quad max\_depth          & UniformInt[1, 10] \\
\quad learning\_rate      & LogUniform[$\exp(-7)$, 1] \\
\quad n\_estimators       & UniformInt[100, 4000] \\
\quad subsample          & UniformFloat[0.2, 1.0] \\
\quad colsample\_bytree   & UniformFloat[0.2, 1.0] \\
\quad min\_child\_weight & LogUniform[$\exp(-16)$, $\exp(5)$] \\
\quad gamma              & LogUniform[$\exp(-16)$, $\exp(2)$] \\
\quad reg\_alpha         & LogUniform[$\exp(-16)$, $\exp(2)$] \\
\quad reg\_lambda        & LogUniform[$\exp(-16)$, $\exp(2)$] \\
\midrule

\multicolumn{2}{l}{\textit{LightGBM}} \\ \midrule
\quad n\_estimators        & UniformInt[50, 10000] \\
\quad learning\_rate       & LogUniform[1e$^{-3}$, 0.3] \\
\quad num\_leaves          & UniformInt[20, 200] \\
\quad max\_depth           & UniformInt[-1, 15] \\
\quad min\_child\_samples  & UniformInt[5, 100] \\
\quad subsample            & UniformFloat[0.2, 1.0] \\
\quad subsample\_freq      & UniformInt[0, 10] \\
\quad colsample\_bytree    & UniformFloat[0.2, 1.0] \\
\quad reg\_alpha           & LogUniform[1e$^{-6}$, 10] \\
\quad reg\_lambda          & LogUniform[1e$^{-6}$, 10] \\
\quad min\_split\_gain     & LogUniform[1e$^{-6}$, 1] \\

\bottomrule
\end{tabular}
}
\end{table}

\newpage
\section{Number of operations and computational cost}\label{App_A}

 \begin{figure}[h]
    \centering
    \includegraphics[width=0.7\linewidth, trim={0.25cm 0 0 0},clip]{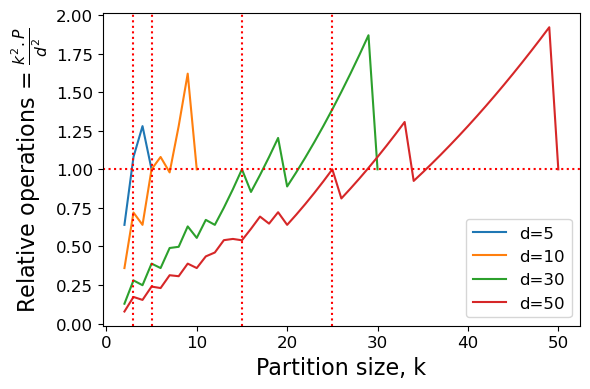}
    \caption{
    Effects of feature partition size (k) on the computational cost of incremental learning compared to learning the complete feature space (d) at a time. The vertical dotted lines mark the feature partition size when its computational cost equals that of learning the complete feature space or the relative operation ratio is 1.0).
    }
    \label{fig:computations-comparison}
\end{figure}

\clearpage

\section{Data set-specific AUC classification scores} \label{AUC_F1}


\begin{table}[h]
\caption{AUC value for MCAR missing type for two missing rates: 10\% and 50\%. Rows are data sets, columns are the models.}
\label{auc-mcar}
\resizebox{\textwidth}{!}{
\begin{tabular}{lllllllllllllllll}
\toprule
Dataset ID             & \% & Median-GBT   & Median-FTT   & MICE-GBT     & MICE-FTT     & GAIN-GBT     & GAIN-FTT     & Diffputer-FTT & Diffputer-GBT & MIA-Xgboost  & MIA-LightGBM & AM-FTT       & IFIAL (K=d/2) \\ 
\midrule
\multirow{2}{*}{13}    & 10 & 0.653 (0.08) & 0.639 (0.09) & 0.661 (0.06) & 0.641 (0.05) & 0.675 (0.08) & 0.654 (0.07) & 0.597 (0.07)  & 0.658 (0.08)  & 0.646 (0.12) & 0.645 (0.08) & 0.611 (0.08) & 0.598 (0.17)  \\
                       & 50 & 0.669 (0.08) & 0.555 (0.11) & 0.648 (0.14) & 0.573 (0.05) & 0.643 (0.10) & 0.602 (0.08) & 0.511 (0.09)  & 0.650 (0.08)  & 0.626 (0.12) & 0.600 (0.13) & 0.538 (0.12) & 0.580 (0.14)  \\
\multirow{2}{*}{31}    & 10 & 0.705 (0.02) & 0.697 (0.03) & 0.714 (0.03) & 0.706 (0.04) & 0.707 (0.02) & 0.683 (0.02) & 0.711 (0.04)  & 0.690 (0.02)  & 0.718 (0.03) & 0.637 (0.02) & 0.696 (0.03) & 0.736 (0.04)  \\
                       & 50 & 0.698 (0.03) & 0.608 (0.06) & 0.689 (0.04) & 0.654 (0.04) & 0.695 (0.04) & 0.643 (0.05) & 0.645 (0.03)  & 0.675 (0.02)  & 0.712 (0.03) & 0.596 (0.03) & 0.606 (0.05) & 0.655 (0.05)  \\
\multirow{2}{*}{37}    & 10 & 0.586 (0.03) & 0.791 (0.04) & 0.589 (0.03) & 0.722 (0.04) & 0.624 (0.04) & 0.753 (0.04) & 0.754 (0.03)  & 0.556 (0.05)  & 0.654 (0.02) & 0.558 (0.04) & 0.754 (0.04) & 0.630 (0.10)  \\
                       & 50 & 0.561 (0.05) & 0.713 (0.06) & 0.593 (0.04) & 0.637 (0.06) & 0.584 (0.02) & 0.598 (0.11) & 0.633 (0.07)  & 0.686 (0.06)  & 0.675 (0.05) & 0.676 (0.07) & 0.653 (0.04) & 0.555 (0.08)  \\
\multirow{2}{*}{181}   & 10 & 0.601 (0.03) & 0.742 (0.02) & 0.645 (0.05) & 0.694 (0.02) & 0.665 (0.05) & 0.758 (0.02) & 0.734 (0.03)  & 0.617 (0.03)  & 0.663 (0.04) & 0.649 (0.04) & 0.688 (0.01) & 0.743 (0.01)  \\
                       & 50 & 0.560 (0.02) & 0.666 (0.03) & 0.590 (0.02) & 0.592 (0.02) & 0.598 (0.04) & 0.655 (0.02) & 0.584 (0.02)  & 0.587 (0.03)  & 0.556 (0.06) & 0.549 (0.00) & 0.592 (0.02) & 0.722 (0.02)  \\
\multirow{2}{*}{334}   & 10 & 0.647 (0.08) & 0.671 (0.15) & 0.673 (0.05) & 0.755 (0.13) & 0.666 (0.08) & 0.906 (0.02) & 0.671 (0.15)  & 0.643 (0.07)  & 0.670 (0.05) & 0.643 (0.07) & 0.772 (0.12) & 0.912 (0.01)  \\
                       & 50 & 0.699 (0.09) & 0.681 (0.04) & 0.670 (0.04) & 0.515 (0.05) & 0.665 (0.07) & 0.527 (0.07) & 0.681 (0.04)  & 0.637 (0.05)  & 0.674 (0.08) & 0.637 (0.05) & 0.503 (0.03) & 0.620 (0.08)  \\
\multirow{2}{*}{463}   & 10 & 0.526 (0.19) & 0.667 (0.04) & 0.679 (0.19) & 0.644 (0.10) & 0.608 (0.09) & 0.546 (0.31) & 0.516 (0.24)  & 0.577 (0.20)  & 0.655 (0.12) & 0.516 (0.24) & 0.673 (0.18) & 0.747 (0.08)  \\
                       & 50 & 0.678 (0.15) & 0.571 (0.04) & 0.670 (0.09) & 0.664 (0.12) & 0.667 (0.14) & 0.567 (0.27) & 0.631 (0.20)  & 0.552 (0.16)  & 0.605 (0.12) & 0.552 (0.16) & 0.537 (0.30) & 0.666 (0.10)  \\
\multirow{2}{*}{1063}  & 10 & 0.699 (0.10) & 0.790 (0.06) & 0.507 (0.05) & 0.788 (0.05) & 0.675 (0.09) & 0.796 (0.05) & 0.797 (0.05)  & 0.697 (0.04)  & 0.686 (0.03) & 0.507 (0.05) & 0.806 (0.05) & 0.771 (0.11)  \\
                       & 50 & 0.793 (0.08) & 0.682 (0.02) & 0.807 (0.04) & 0.783 (0.05) & 0.790 (0.04) & 0.786 (0.05) & 0.799 (0.05)  & 0.821 (0.05)  & 0.845 (0.03) & 0.682 (0.02) & 0.810 (0.05) & 0.805 (0.05)  \\
\multirow{2}{*}{1067}  & 10 & 0.769 (0.02) & 0.741 (0.02) & 0.936 (0.01) & 0.743 (0.02) & 0.897 (0.02) & 0.746 (0.02) & 0.757 (0.02)  & 0.937 (0.01)  & 0.936 (0.01) & 0.741 (0.02) & 0.747 (0.03) & 0.808 (0.03)  \\
                       & 50 & 0.699 (0.03) & 0.645 (0.04) & 0.750 (0.04) & 0.731 (0.02) & 0.736 (0.03) & 0.738 (0.01) & 0.735 (0.02)  & 0.757 (0.03)  & 0.699 (0.03) & 0.645 (0.04) & 0.674 (0.02) & 0.794 (0.02)  \\
\multirow{2}{*}{1071}  & 10 & 0.651 (0.09) & 0.632 (0.22) & 0.634 (0.13) & 0.728 (0.09) & 0.662 (0.16) & 0.745 (0.07) & 0.684 (0.11)  & 0.651 (0.09)  & 0.634 (0.13) & 0.553 (0.11) & 0.662 (0.16) & 0.755 (0.09)  \\
                       & 50 & 0.669 (0.10) & 0.615 (0.23) & 0.663 (0.13) & 0.696 (0.12) & 0.617 (0.17) & 0.717 (0.10) & 0.691 (0.10)  & 0.638 (0.14)  & 0.617 (0.17) & 0.607 (0.14) & 0.617 (0.17) & 0.648 (0.10)  \\
\multirow{2}{*}{1480}  & 10 & 0.662 (0.05) & 0.658 (0.05) & 0.626 (0.03) & 0.692 (0.05) & 0.658 (0.05) & 0.674 (0.06) & 0.673 (0.08)  & 0.655 (0.07)  & 0.626 (0.03) & 0.613 (0.04) & 0.613 (0.04) & 0.707 (0.07)  \\
                       & 50 & 0.503 (0.09) & 0.570 (0.10) & 0.652 (0.05) & 0.646 (0.02) & 0.546 (0.06) & 0.600 (0.06) & 0.591 (0.06)  & 0.627 (0.05)  & 0.546 (0.06) & 0.503 (0.09) & 0.546 (0.06) & 0.715 (0.09)  \\
\multirow{2}{*}{1498}  & 10 & 0.642 (0.06) & 0.705 (0.03) & 0.655 (0.06) & 0.694 (0.05) & 0.683 (0.05) & 0.695 (0.06) & 0.688 (0.03)  & 0.639 (0.05)  & 0.642 (0.06) & 0.639 (0.07) & 0.683 (0.05) & 0.696 (0.07)  \\
                       & 50 & 0.646 (0.07) & 0.643 (0.03) & 0.670 (0.04) & 0.681 (0.04) & 0.604 (0.10) & 0.657 (0.03) & 0.599 (0.06)  & 0.472 (0.07)  & 0.599 (0.06) & 0.546 (0.06) & 0.604 (0.10) & 0.678 (0.03)  \\
\multirow{2}{*}{23381} & 10 & 0.551 (0.06) & 0.532 (0.03) & 0.564 (0.02) & 0.532 (0.03) & 0.543 (0.07) & 0.526 (0.10) & 0.543 (0.07)  & 0.572 (0.05)  & 0.532 (0.03) & 0.526 (0.10) & 0.553 (0.02) & 0.612 (0.03)  \\
                       & 50 & 0.560 (0.06) & 0.638 (0.11) & 0.565 (0.05) & 0.638 (0.11) & 0.564 (0.03) & 0.511 (0.05) & 0.507 (0.05)  & 0.564 (0.03)  & 0.511 (0.05) & 0.507 (0.05) & 0.626 (0.07) & 0.555 (0.05)  \\
\multirow{2}{*}{40691} & 10 & 0.627 (0.01) & 0.696 (0.01) & 0.613 (0.04) & 0.654 (0.05) & 0.683 (0.02) & 0.685 (0.02) & 0.683 (0.02)  & 0.582 (0.03)  & 0.683 (0.02) & 0.654 (0.05) & 0.674 (0.03) & 0.722 (0.01)  \\
                       & 50 & 0.594 (0.05) & 0.619 (0.01) & 0.568 (0.02) & 0.594 (0.02) & 0.604 (0.02) & 0.632 (0.02) & 0.554 (0.03)  & 0.562 (0.04)  & 0.594 (0.02) & 0.554 (0.03) & 0.604 (0.02) & 0.687 (0.02)  \\
                       \bottomrule
\end{tabular}}
\end{table}

\begin{table}[h]
\caption{AUC value for MNAR missing type for two missing rates: 10\% and 50\%. Rows are data sets, columns are the models.}
\label{auc-mnar}
\resizebox{\textwidth}{!}{
\begin{tabular}{lllllllllllllllll}
\toprule
Dataset ID & \% & Median-GBT   & Median-FTT   & MICE-GBT     & MICE-FTT     & GAIN-GBT     & GAIN-FTT     & Diffputer-GBT & Diffputer-FTT & MIA-Xgboost  & MIA-LightGBM & AM-FTT       & IFIAL (K=d/2) \\ 
\midrule
\multirow{2}{*}{13}    & 10 & 0.704 (0.08) & 0.660 (0.08) & 0.704 (0.08) & 0.658 (0.06) & 0.701 (0.09) & 0.698 (0.08) & 0.667 (0.08)  & 0.694 (0.06)  & 0.550 (0.13) & 0.650 (0.06) & 0.660 (0.08) & 0.675 (0.06)  \\
                       & 50 & 0.642 (0.09) & 0.606 (0.08) & 0.642 (0.09) & 0.638 (0.10) & 0.676 (0.10) & 0.617 (0.10) & 0.628 (0.07)  & 0.674 (0.09)  & 0.550 (0.16) & 0.633 (0.11) & 0.606 (0.08) & 0.637 (0.11)  \\
\multirow{2}{*}{31}    & 10 & 0.699 (0.01) & 0.745 (0.04) & 0.695 (0.01) & 0.687 (0.04) & 0.730 (0.03) & 0.697 (0.03) & 0.710 (0.04)  & 0.739 (0.04)  & 0.582 (0.11) & 0.580 (0.06) & 0.550 (0.02) & 0.708 (0.05)  \\
                       & 50 & 0.641 (0.02) & 0.679 (0.01) & 0.635 (0.03) & 0.644 (0.03) & 0.657 (0.03) & 0.661 (0.03) & 0.618 (0.06)  & 0.645 (0.04)  & 0.550 (0.02) & 0.573 (0.06) & 0.567 (0.03) & 0.650 (0.06)  \\
\multirow{2}{*}{37}    & 10 & 0.565 (0.02) & 0.578 (0.03) & 0.550 (0.05) & 0.582 (0.02) & 0.555 (0.01) & 0.590 (0.02) & 0.601 (0.05)  & 0.575 (0.02)  & 0.559 (0.01) & 0.577 (0.04) & 0.556 (0.03) & 0.805 (0.04)  \\
                       & 50 & 0.559 (0.05) & 0.551 (0.07) & 0.571 (0.05) & 0.570 (0.03) & 0.558 (0.02) & 0.558 (0.04) & 0.576 (0.05)  & 0.550 (0.05)  & 0.556 (0.01) & 0.562 (0.04) & 0.553 (0.04) & 0.622 (0.03)  \\
\multirow{2}{*}{181}   & 10 & 0.568 (0.03) & 0.560 (0.01) & 0.573 (0.03) & 0.605 (0.04) & 0.550 (0.04) & 0.600 (0.03) & 0.556 (0.05)  & 0.560 (0.01)  & 0.552 (0.02) & 0.565 (0.02) & 0.608 (0.01) & 0.742 (0.02)  \\
                       & 50 & 0.550 (0.02) & 0.566 (0.02) & 0.559 (0.02) & 0.572 (0.02) & 0.563 (0.05) & 0.576 (0.03) & 0.565 (0.03)  & 0.561 (0.02)  & 0.568 (0.01) & 0.566 (0.02) & 0.573 (0.01) & 0.617 (0.02)  \\
\multirow{2}{*}{334}   & 10 & 0.661 (0.05) & 0.806 (0.10) & 0.661 (0.05) & 0.767 (0.04) & 0.615 (0.05) & 0.811 (0.04) & 0.643 (0.06)  & 0.742 (0.17)  & 0.550 (0.04) & 0.767 (0.11) & 0.806 (0.10) & 0.744 (0.10)  \\
                       & 50 & 0.558 (0.01) & 0.594 (0.05) & 0.558 (0.01) & 0.582 (0.05) & 0.555 (0.06) & 0.595 (0.02) & 0.550 (0.04)  & 0.558 (0.04)  & 0.597 (0.03) & 0.573 (0.05) & 0.594 (0.05) & 0.616 (0.05)  \\
\multirow{2}{*}{463}   & 10 & 0.627 (0.09) & 0.675 (0.06) & 0.596 (0.11) & 0.595 (0.10) & 0.672 (0.10) & 0.665 (0.08) & 0.556 (0.08)  & 0.556 (0.21)  & 0.627 (0.22) & 0.585 (0.12) & 0.550 (0.09) & 0.557 (0.19)  \\
                       & 50 & 0.653 (0.13) & 0.601 (0.12) & 0.622 (0.14) & 0.554 (0.11) & 0.566 (0.09) & 0.627 (0.10) & 0.663 (0.09)  & 0.572 (0.11)  & 0.601 (0.15) & 0.550 (0.12) & 0.607 (0.21) & 0.488 (0.06)  \\
\multirow{2}{*}{1063}  & 10 & 0.892 (0.05) & 0.671 (0.11) & 0.863 (0.04) & 0.912 (0.04) & 0.572 (0.03) & 0.904 (0.04) & 0.550 (0.06)  & 0.595 (0.11)  & 0.582 (0.02) & 0.707 (0.14) & 0.797 (0.07) & 0.846 (0.06)  \\
                       & 50 & 0.597 (0.06) & 0.585 (0.05) & 0.564 (0.04) & 0.590 (0.08) & 0.563 (0.02) & 0.588 (0.07) & 0.565 (0.08)  & 0.550 (0.05)  & 0.568 (0.03) & 0.559 (0.05) & 0.554 (0.04) & 0.705 (0.13)  \\
\multirow{2}{*}{1067}  & 10 & 0.802 (0.02) & 0.711 (0.04) & 0.987 (0.01) & 0.993 (0.01) & 0.579 (0.01) & 0.992 (0.01) & 0.567 (0.01)  & 0.798 (0.11)  & 0.550 (0.09) & 0.957 (0.01) & 0.978 (0.01) & 0.762 (0.04)  \\
                       & 50 & 0.712 (0.08) & 0.636 (0.02) & 0.796 (0.03) & 0.801 (0.03) & 0.569 (0.01) & 0.796 (0.04) & 0.550 (0.03)  & 0.613 (0.04)  & 0.599 (0.09) & 0.735 (0.06) & 0.792 (0.02) & 0.743 (0.02)  \\
\multirow{2}{*}{1071}  & 10 & 0.625 (0.08) & 0.550 (0.10) & 0.589 (0.06) & 0.602 (0.07) & 0.589 (0.03) & 0.595 (0.07) & 0.616 (0.04)  & 0.589 (0.09)  & 0.624 (0.03) & 0.613 (0.10) & 0.552 (0.12) & 0.690 (0.13)  \\
                       & 50 & 0.612 (0.03) & 0.560 (0.11) & 0.573 (0.12) & 0.572 (0.10) & 0.553 (0.01) & 0.584 (0.11) & 0.574 (0.08)  & 0.551 (0.06)  & 0.550 (0.01) & 0.571 (0.12) & 0.582 (0.07) & 0.603 (0.13)  \\
\multirow{2}{*}{1480}  & 10 & 0.565 (0.05) & 0.561 (0.04) & 0.564 (0.04) & 0.569 (0.04) & 0.568 (0.02) & 0.573 (0.03) & 0.584 (0.05)  & 0.561 (0.03)  & 0.550 (0.05) & 0.554 (0.04) & 0.553 (0.07) & 0.587 (0.02)  \\
                       & 50 & 0.553 (0.07) & 0.554 (0.04) & 0.552 (0.03) & 0.551 (0.08) & 0.556 (0.03) & 0.550 (0.08) & 0.554 (0.05)  & 0.552 (0.06)  & 0.552 (0.04) & 0.551 (0.04) & 0.553 (0.04) & 0.556 (0.01)  \\
\multirow{2}{*}{1498}  & 10 & 0.632 (0.09) & 0.642 (0.05) & 0.617 (0.09) & 0.604 (0.06) & 0.642 (0.06) & 0.588 (0.05) & 0.604 (0.05)  & 0.632 (0.06)  & 0.630 (0.10) & 0.550 (0.03) & 0.567 (0.04) & 0.657 (0.04)  \\
                       & 50 & 0.578 (0.06) & 0.598 (0.05) & 0.568 (0.03) & 0.550 (0.08) & 0.583 (0.02) & 0.554 (0.09) & 0.582 (0.05)  & 0.590 (0.03)  & 0.569 (0.06) & 0.431 (0.06) & 0.559 (0.07) & 0.503 (0.07)  \\
\multirow{2}{*}{23381} & 10 & 0.582 (0.06) & 0.628 (0.06) & 0.582 (0.06) & 0.592 (0.05) & 0.572 (0.04) & 0.570 (0.05) & 0.550 (0.16)  & 0.625 (0.06)  & 0.556 (0.08) & 0.596 (0.02) & 0.572 (0.10) & 0.642 (0.04)  \\
                       & 50 & 0.595 (0.06) & 0.590 (0.05) & 0.597 (0.07) & 0.550 (0.04) & 0.586 (0.13) & 0.604 (0.03) & 0.570 (0.15)  & 0.574 (0.02)  & 0.606 (0.11) & 0.554 (0.01) & 0.578 (0.06) & 0.611 (0.02)  \\
\multirow{2}{*}{40691} & 10 & 0.595 (0.04) & 0.550 (0.03) & 0.551 (0.06) & 0.585 (0.05) & 0.560 (0.03) & 0.583 (0.04) & 0.559 (0.02)  & 0.560 (0.03)  & 0.564 (0.01) & 0.570 (0.02) & 0.713 (0.01) & 0.670 (0.03)  \\
                       & 50 & 0.571 (0.05) & 0.568 (0.02) & 0.571 (0.01) & 0.559 (0.05) & 0.565 (0.02) & 0.550 (0.06) & 0.559 (0.03)  & 0.564 (0.03)  & 0.557 (0.01) & 0.567 (0.03) & 0.644 (0.02) & 0.647 (0.01)  \\
                       \bottomrule
\end{tabular}}
\end{table}

\end{document}